\PassOptionsToPackage{unicode}{hyperref}
\PassOptionsToPackage{hyphens}{url}
\PassOptionsToPackage{dvipsnames,svgnames,x11names}{xcolor}
\documentclass[
  11pt,
]{article}
\usepackage{xcolor}
\usepackage[margin=1in]{geometry}
\usepackage{amsmath,amssymb}
\usepackage{graphicx}
\usepackage{caption}
\usepackage{booktabs}
\usepackage{iftex}
\ifPDFTeX
  \usepackage[T1]{fontenc}
  \usepackage[utf8]{inputenc}
  \usepackage{textcomp} 
\else 
  \usepackage{unicode-math} 
  \defaultfontfeatures{Scale=MatchLowercase}
  \defaultfontfeatures[\rmfamily]{Ligatures=TeX,Scale=1}
\fi
\usepackage{lmodern}
\ifPDFTeX\else
\fi
\IfFileExists{upquote.sty}{\usepackage{upquote}}{}
\IfFileExists{microtype.sty}{
  \usepackage[]{microtype}
  \UseMicrotypeSet[protrusion]{basicmath} 
}{}
\makeatletter
\@ifundefined{KOMAClassName}{
  \IfFileExists{parskip.sty}{%
    \usepackage{parskip}
  }{
    \setlength{\parindent}{0pt}
    \setlength{\parskip}{6pt plus 2pt minus 1pt}}
}{
  \KOMAoptions{parskip=half}}
\makeatother
\providecommand{\tightlist}{%
  \setlength{\itemsep}{0pt}\setlength{\parskip}{0pt}}
\usepackage{bookmark}
\IfFileExists{xurl.sty}{\usepackage{xurl}}{} 
\hypersetup{
  pdftitle={Beyond LLM Wrappers: Context Engineering and the Three-Engine Architecture for Reliable Financial AI},
  pdfauthor={Pradeep Kumar GS, Finteract AI; Ravi Tanniru, Finteract AI},
  colorlinks=true,
  linkcolor={blue},
  filecolor={Maroon},
  citecolor={blue},
  urlcolor={blue},
  pdfcreator={LaTeX via pandoc}}

\title{Beyond LLM Wrappers: Context Engineering and the Three-Engine
Architecture for Reliable Financial AI}
\author{Pradeep Kumar GS\\
  Finteract AI\\
  \texttt{pradeep.kumar@finteract.ai}
  \and
  Ravi Tanniru\\
  Finteract AI\\
  \texttt{ravi.tanniru@finteract.ai}}
\date{June 2026}

\begin{document}
\maketitle

\section{Abstract}\label{abstract}

Financial AI systems are increasingly expected to answer operational
questions that are not merely informational, but decision-bearing: which
vendors should be paid this week, which customers should be escalated
for collection, which invoices are exposed to tax risk, what cash
position is likely under a delayed-receivables scenario, and why a
working-capital metric changed. These questions require natural language
interaction, but they also require deterministic computation, persistent
business context, source traceability, and auditable recommendations. A
large language model (LLM) by itself cannot satisfy those requirements.
It can interpret language and produce explanations, but it does not
provide a reliable memory of financial relationships, does not guarantee
reproducible calculations, and can generate plausible but unsupported
claims. Conversely, knowledge graphs, optimization solvers, and rules
engines each provide important capabilities, but none provides the
complete user-facing experience required by finance teams.

This paper presents a production-grounded architecture for reliable
financial AI based on three engines: a financial knowledge graph, a
deterministic optimization engine, and a multi-agent LLM layer. The
knowledge graph stores enterprise financial context, relationships,
provenance, and temporal state. The optimization engine performs
reproducible calculations for payment scheduling, cash allocation,
vendor prioritization, discount capture, and similar decision tasks. The
multi-agent LLM layer interprets user intent, formulates tasks,
assembles explanations, and validates generated claims against
structured ground truth. The central contribution is not merely the use
of these engines, but the coordination layer that connects them. We call
this layer context engineering: the process of selecting, ranking,
composing, validating, and budgeting the context required for each
financial query across heterogeneous data sources, tools, tenants, and
decision objectives.

Drawing from 15 production deployments over approximately 11 months,
this paper reports several observations. First, single-agent LLM pipelines produced
unacceptable hallucination rates (approximately 7--9\%, ranging 5--10\% by
query type) for financial recommendations, including fabricated
invoices, invented discounts, incorrect totals, and unsupported vendor
claims. Second, graph-backed multi-agent validation reduced observed
hallucinations to below 1\% in production usage, although the evidence
is observational rather than randomized. Third, ERP data quality and
tenant heterogeneity dominated implementation effort: inconsistent
vendor names, missing fields, duplicate entities, free-text terms, and
industry-specific schemas required substantial normalization before AI
reasoning could be trusted. Fourth, deployment time fell from roughly
six weeks to approximately two weeks as reusable ERP mappings, graph
templates, validation patterns, and compliance-specific context policies
accumulated.

The paper argues that financial context engineering is distinct from
standard retrieval-augmented generation. Financial systems must retrieve
structured relationships rather than only text chunks, respect temporal
validity, include exact numerical evidence, enforce access control,
adapt to tenant-specific schemas, and validate claims against source
records and solver outputs. We position context engineering as a
research problem for high-stakes structured domains and identify open
problems in evaluation, reproducibility, privacy, benchmark design,
context quality metrics, and cross-tenant pattern learning.

\textbf{Keywords:} financial AI, context engineering, knowledge graphs,
multi-agent systems, deterministic optimization, audit trails,
retrieval-augmented generation, working capital, ERP integration

\section{1. Introduction}\label{introduction}

Finance work is full of questions that look simple in natural language
and complex in execution. A controller may ask, ``Which vendors should
we prioritize for payment this week?'' A CFO may ask, ``What happens to
cash if our largest customer delays payment by ten days?'' An accounts
payable manager may ask, ``Why did aging in the 60-day bucket increase
this month?'' A tax reviewer may ask, ``Which invoices are exposed to
GST mismatch risk?'' Each question requires a fluent answer, but the
answer is only useful if it is grounded in source records, reflects
current financial state, respects business constraints, and can be
audited after the decision is made.

This is the core difficulty for financial AI. The interface problem and
the reasoning problem are different. Natural language is the desired
interface because finance users should not need to write SQL, Cypher,
optimization code, or ERP-specific queries. Yet the underlying decision
must be derived from invoices, payments, contracts, tax codes, bank
balances, purchase orders, customer terms, vendor relationships, and
historical behavior. The system must map a conversational question to
structured evidence, perform deterministic calculations where necessary,
explain the result in business terms, and preserve a trace that another
person can inspect.

LLMs are strong at language understanding, summarization, and
interactive explanation, but financial AI cannot be built as a thin
wrapper around an LLM. The discount must actually exist. The cash
position must come from a source system. Calculations must be
reproducible. In financial systems, a hallucination has operational
shape---a nonexistent invoice number, an invented payment term, a
misidentified vendor, or a recommendation that violates a cash
constraint---and a response that is 95\% correct may still be
unacceptable if the remaining 5\% contains the vendor, amount, or date
that drives the decision. Traditional deterministic systems---ERP
reports, rules engines, optimization solvers---face the opposite
limitation: exact but rigid, requiring users to know where to look and
how to interpret exceptions. The problem is not choosing one engine. The
problem is composing engines so that each performs the role for which it
is suited.

This paper proposes a three-engine architecture for reliable financial
AI. The first engine is a financial knowledge graph. It functions as
enterprise memory: a structured, temporal, provenance-bearing
representation of financial entities and relationships. The second
engine is a deterministic optimization layer. It performs calculations
and constrained decision-making for payment schedules, cash allocation,
discount capture, collections prioritization, and similar tasks. The
third engine is a multi-agent LLM layer. It handles language, intent,
task formulation, explanation, and validation orchestration. These
engines are coordinated by context engineering, a layer that determines
what evidence, graph substructures, solver outputs, tenant settings, and
validation checks should be assembled for each query.

The term context engineering is used deliberately. Standard
retrieval-augmented generation (RAG) generally retrieves text chunks
that are semantically similar to a question and supplies them to a
model. Financial AI needs more than similarity. It needs structured
traversal, entity resolution, temporal filtering, context budgeting,
validation, tenant-specific interpretation, and deterministic tool
invocation. If a user asks whether to pay Vendor A now or next week, the
relevant context may include unpaid invoices, cash balances, discount
windows, vendor criticality, tax obligations, purchase order status,
historical payment behavior, and policy constraints. Some of that
context is text, some is structured data, some is computed output, and
some is a validation trace. Context engineering is the process of
turning that heterogeneous evidence into the right model input and the
right audit output.

The production motivation for this work comes from 15 production
deployments spanning approximately 11 months. The deployed settings span
manufacturing, SaaS, services, e-commerce, and trading businesses, with
ERP integrations including Zoho, Tally, QuickBooks, and Acumatica. These
environments are messy in ways that matter: vendor names are
inconsistent, payment terms are free text, source records are
incomplete, tenant definitions vary by industry, and financial workflows
mix deterministic rules with judgment. These deployments showed that the
most difficult part of financial AI was not prompting a model. It was
constructing enough trusted context for the model, solver, and graph to
work together.

The framing is intentionally operational rather than speculative. Many
discussions of AI in finance focus on model capability: whether an LLM
can read a statement, summarize a filing, or answer questions about a
policy document. Those tasks matter, but they do not capture the
pressure of day-to-day finance operations. Operational finance decisions
are time-sensitive, cross-system, and consequence-bearing. A delayed
payment can damage a supplier relationship. A missed discount can create
measurable cash leakage. A wrong tax classification can trigger review.
A stale receivables assumption can distort a cash forecast. For these
tasks, the system must know not only what words mean, but what records
are current, which constraints bind, which exceptions matter, and how to
justify a recommendation after the fact.

This paper therefore treats financial AI as decision infrastructure. The
unit of analysis is not a prompt-response pair. It is the full path from
user intent to source records, context assembly, deterministic
computation, validation, explanation, and audit. That path determines
whether an AI system is usable by finance teams. A model may be
powerful, but if it receives stale context, it will produce stale
answers. A graph may be complete, but if the system retrieves the wrong
subgraph, the answer will omit critical constraints. A solver may be
exact, but if the LLM formulates the wrong objective, the result will
optimize the wrong business problem. Context engineering is the
discipline that makes these dependencies explicit.

Section 2 analyzes the requirements for reliable financial AI. Section 3
describes the three-engine architecture. Section 4 defines context
engineering and explains why it differs from standard RAG. Section 5
discusses integration challenges. Section 6 summarizes production
deployment observations. Section 7 positions the work relative to LLMs,
knowledge graphs, optimization, multi-agent systems, and RAG. Section 8
presents limitations. Section 9 concludes with open research directions.

\section{2. Requirements Analysis}\label{requirements-analysis}

Financial AI systems must satisfy four requirements simultaneously. The
requirements are individually familiar, but their combination is what
makes the domain difficult. A system that satisfies only one or two of
them may be useful as a report generator, chatbot, or analytics
assistant, but not as a trustworthy decision-support system for finance
operations.

The first requirement is contextual reasoning over relationships.
Financial decisions rarely depend on isolated rows. A vendor payment
decision may depend on the vendor's unpaid invoices, payment history,
early-payment discounts, purchase order status, supplier criticality,
contract clauses, parent-subsidiary relationships, cash position, tax
exposure, and operational impact. A collections decision may depend on
customer aging, communication history, promised payment dates, service
delivery status, credit terms, dispute status, and relationship value. A
tax risk decision may depend on invoice classification, jurisdiction,
filing period, document completeness, and reconciliation against ERP
records. These are graph-shaped problems even when the source data lives
in tables.

Relational databases can store these facts, but relationship reasoning
requires the system to know which relationships matter for each query. A
user asking ``Why is Vendor X blocked?'' may need a path from vendor to
unpaid invoice to purchase order to goods receipt to dispute record. A
user asking ``Why did cash conversion worsen?'' may need paths across
receivables, inventory, payables, and revenue recognition. A user asking
``Which customers should we call today?'' may need customer hierarchy,
overdue exposure, promised payments, and strategic account flags. A flat
text retrieval system may find documents that mention the right vendor,
but it does not necessarily assemble the relationship path that explains
the decision.

The second requirement is deterministic computation. Many finance
questions require exact arithmetic or constrained optimization. Payment
scheduling requires deciding which invoices to pay under cash limits,
due dates, discounts, vendor criticality, and policy constraints. Cash
allocation requires choosing how to distribute limited funds across
obligations. Working-capital planning requires projecting cash inflows
and outflows under assumptions. Tax review requires applying rules to
amounts, dates, and jurisdictions. These tasks cannot be delegated to a
probabilistic text generator as the final computational authority. Given
the same inputs, solver formulation, and data snapshot, the result
should be reproducible.

Determinism matters for trust, regression testing, audit, and
operational repeatability. A CFO cannot approve a payment recommendation
if rerunning the same query produces a different schedule without a data
change. A controller cannot investigate a disputed recommendation if the
system cannot reconstruct the exact calculation. Engineers cannot test
financial AI reliably if outputs vary because arithmetic is embedded in
model generation rather than executed by deterministic components. This
does not mean every part of the system must be deterministic. Natural
language interpretation can involve probabilistic models. But the
numerical core of financial decisions must be handled by deterministic
tools.

The third requirement is natural language interaction. Finance teams do
not want to operate AI systems by writing graph queries, solver
formulations, or ERP-specific scripts. They want to ask business
questions. A natural language interface is not a cosmetic layer; it
changes who can use advanced analysis. The accounts payable manager
should be able to ask, ``What should we pay today if we hold back 20\%
cash for payroll?'' The CFO should be able to ask, ``What are the top
three drivers of the cash shortfall next week?'' A tax analyst should be
able to ask, ``Which invoices need review before filing?'' The system
must translate these questions into the right combination of retrieval,
computation, validation, and explanation.

Natural language also matters for explanations. Finance decisions are
collaborative. Recommendations must be presented in language that users
can challenge: ``Vendor A is prioritized because three invoices totaling
48,000 are due this week, a 2\% discount expires tomorrow, and the
vendor supplies a critical component. Vendor B is deferred because its
invoices are not yet due and no discount is active.'' Such explanations
are not just summaries. They are the bridge between computation and
decision accountability.

The fourth requirement is auditability. Financial systems must preserve
data lineage, process provenance, decision rationale, and temporal
validity. Data lineage answers where a fact came from: ERP source,
record identifier, synchronization timestamp, transformation version,
and entity-resolution rule. Process provenance answers how the system
moved from user query to context retrieval, solver invocation,
validation, and response. Decision rationale explains why the
recommendation follows from evidence and constraints. Temporal validity
ensures that the answer was correct for the relevant time window and
data snapshot.

Auditability is not optional in finance. It is needed for compliance,
internal controls, user trust, dispute resolution, and model governance.
If a system recommends delaying a payment, the organization may need to
explain the recommendation later. If it identifies a tax exposure, a
reviewer may need to inspect the supporting records. If it prioritizes
one vendor over another, procurement and operations teams may need to
understand the trade-off. A financial AI answer without an audit trail
is a liability.

The key observation is that current engine types satisfy these
requirements only partially. LLMs provide language understanding and
explanation, but are probabilistic and lack a native persistent
representation of enterprise state. Knowledge graphs provide structured
context and relationship traversal, but do not by themselves optimize
decisions or converse naturally. Optimization engines provide
deterministic outputs, but require precise inputs and cannot infer
business intent from ordinary language. Rules-based systems provide
repeatability and explicit logic, but become brittle as tenants,
industries, compliance regimes, and exceptions multiply.

This requirement analysis explains why single-engine approaches fail. An
LLM-only assistant can answer fluently but may hallucinate facts or
calculations. A graph-only system can retrieve relationships but may
require technical queries and lacks optimization. A solver-only workflow
can compute a schedule but cannot discover the right context or explain
the answer conversationally. A rules-only workflow can automate known
cases but struggles with novel combinations of constraints and business
context. Reliable financial AI requires a composition of engines, each
constrained to the tasks it can perform reliably.

The requirements also create design principles. First, source truth must
live outside the LLM in structured systems. Second, arithmetic and
optimization must be delegated to deterministic engines. Third, LLMs
should be used for interpretation, orchestration, and explanation, not
as the source of financial facts. Fourth, every answer should be
validatable against graph records, solver outputs, and logs. Fifth,
context assembly should be treated as a first-class engineering problem
rather than an incidental prompt construction step.

These principles lead directly to the three-engine architecture
described in the next section.

\begin{table}[htbp]
\centering
\caption{Capability matrix: no single engine type satisfies all financial AI requirements simultaneously. The three-engine architecture combines the complementary strengths of all three.}
\label{tab:capability-matrix}
\begin{tabular}{lcccc}
\toprule
\textbf{Engine / System} & \textbf{Contextual} & \textbf{Deterministic} & \textbf{Persistent} & \textbf{Natural} \\
                         & \textbf{Reasoning}  & \textbf{Math}          & \textbf{Context}    & \textbf{Language} \\
\midrule
LLM alone                & $\checkmark$ & $\times$     & $\times$     & $\checkmark$ \\
Knowledge graph alone    & $\times$     & $\times$     & $\checkmark$ & $\times$     \\
Optimization solver alone& $\times$     & $\checkmark$ & $\times$     & $\times$     \\
Rules-based system       & $\times$     & $\checkmark$ & $\times$     & $\times$     \\
\midrule
\textbf{Three-engine architecture} & $\checkmark$ & $\checkmark$ & $\checkmark$ & $\checkmark$ \\
\bottomrule
\end{tabular}
\end{table}

\section{3. Three-Engine Architecture}\label{three-engine-architecture}

\begin{figure}[htbp]
\centering
\includegraphics[width=\linewidth]{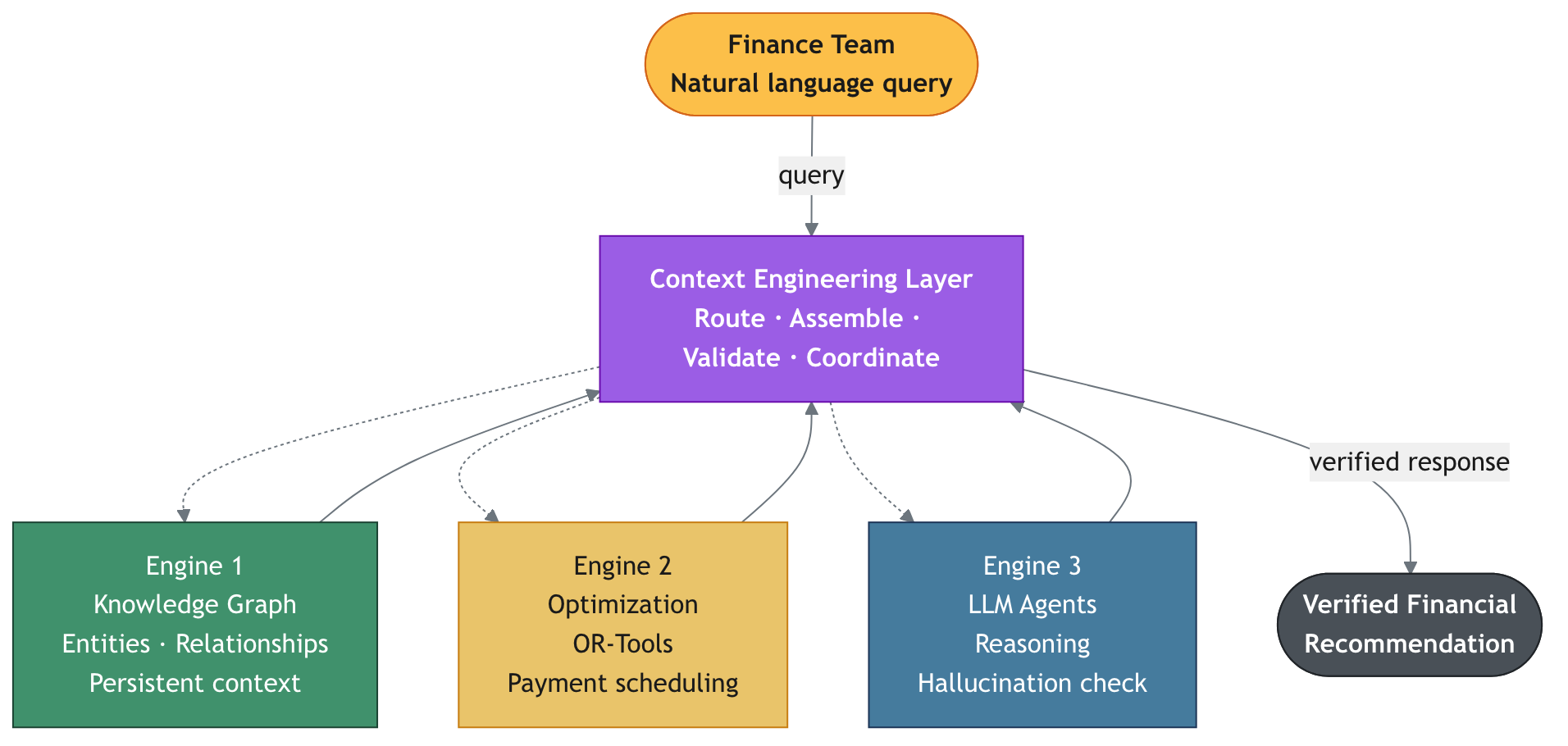}
\caption{Three-engine architecture for reliable financial AI. Knowledge graphs provide persistent context, optimization engines provide deterministic computation, and LLM agents provide query understanding and explanation. The context engineering layer coordinates engine invocation, context assembly, validation, and response composition.}
\label{fig:three-engine}
\end{figure}

The proposed architecture combines three specialized engines with a
coordination layer. The engines are not interchangeable. Each has a
distinct role that maps to the requirements in Section 2. The knowledge
graph provides contextual reasoning and audit lineage. The optimization
engine provides deterministic calculation. The multi-agent LLM layer
provides natural language interaction, task formulation, explanation,
and validation orchestration. Context engineering coordinates the
engines.

The first engine is a financial knowledge graph. Its purpose is to
represent the enterprise's financial world as connected entities and
relationships. Typical nodes include organizations, legal entities,
business units, vendors, customers, invoices, payments, purchase orders,
contracts, tax obligations, bank accounts, journal entries, and users.
Typical relationships include ``issued by,'' ``owed to,'' ``paid by,''
``matched to,'' ``governed by,'' ``belongs to entity,'' ``has tax
code,'' ``depends on,'' and ``part of customer group.'' Each node and
relationship carries properties, timestamps, source metadata, and
transformation history.

The graph provides three capabilities that are difficult to obtain from
isolated tables or documents. First, it supports multi-hop reasoning. A
question about a payment can traverse from vendor to invoice to purchase
order to goods receipt to dispute status. Second, it supports temporal
context. Historical invoice states, aging transitions, payment behavior,
and synchronization timestamps can be represented and queried. Third, it
supports source attribution. A retrieved fact can carry ERP system,
source record ID, sync time, version, and normalization rule.

\begin{figure}[htbp]
\centering
\includegraphics[width=\linewidth]{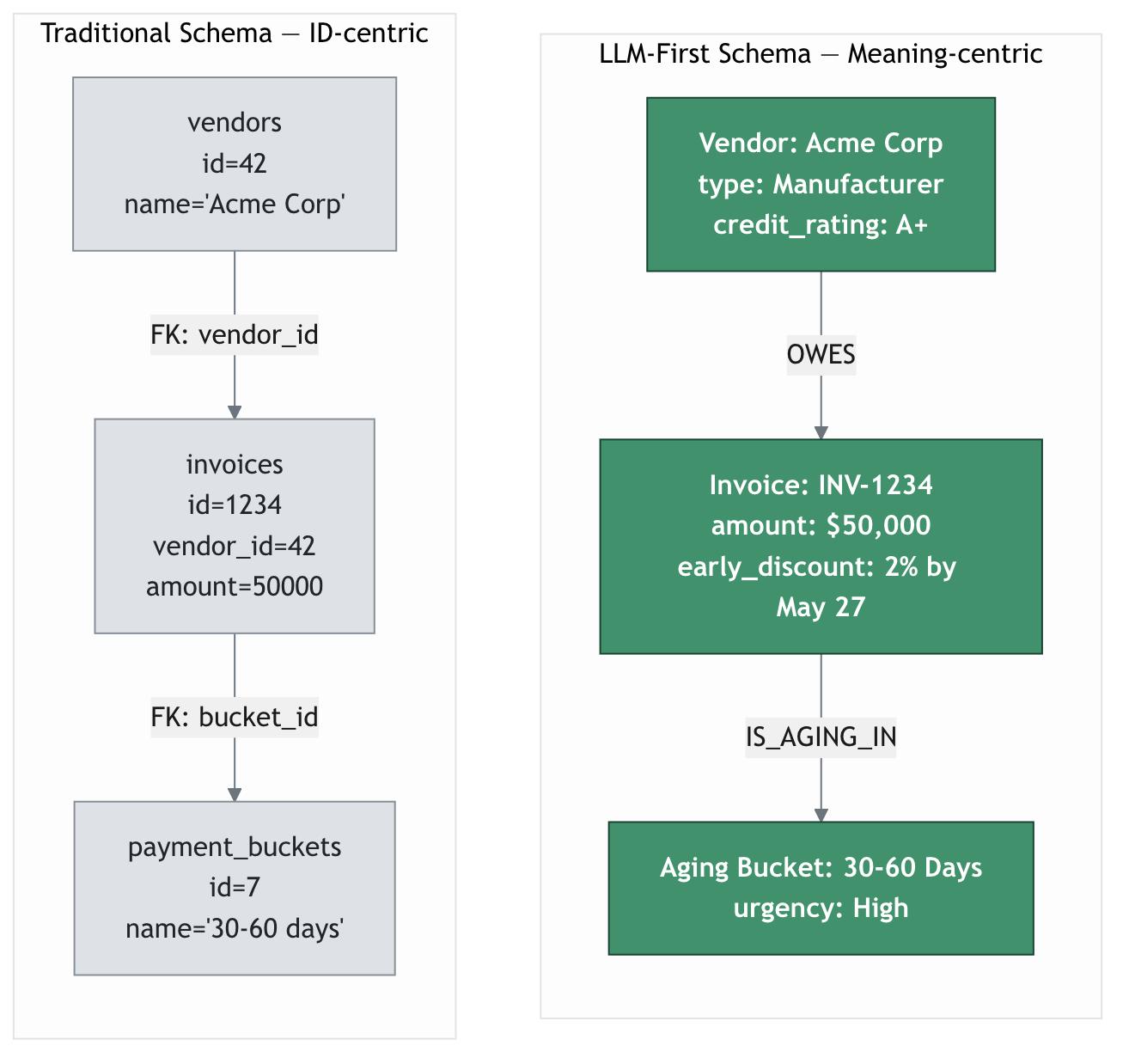}
\caption{LLM-first knowledge graph schema principles for financial AI: descriptive properties, hierarchical structures, and business-meaningful relationships.}
\label{fig:kg-schema}
\end{figure}

Graph design must be LLM-aware. Traditional enterprise schemas often
optimize for storage efficiency, transaction processing, or formal
ontology coverage. An LLM-facing graph needs property names,
relationship labels, summaries, and hierarchy that are readable by
agents. For example, a relationship called \texttt{HAS\_OPEN\_INVOICE}
is more useful for context assembly than a cryptic join table name. A
vendor node that includes normalized name, aliases, criticality, payment
behavior summary, and source identifiers is more useful than a sparse
record that requires many additional joins. The graph is not simply a
database; it is a context substrate for reasoning agents.

The second engine is a deterministic optimization layer. Financial
operations contain many constrained decisions. A payment schedule may
maximize discount capture and vendor reliability subject to cash
availability, due dates, payroll reserve, tax obligations, strategic
supplier constraints, and policy rules. A collections prioritization
model may rank customers by overdue amount, likelihood of recovery,
account importance, dispute status, and collector capacity. A cash-flow
scenario may allocate expected inflows to obligations while preserving
minimum liquidity. These are not text-generation tasks. They are
optimization tasks.

The optimization engine accepts structured inputs from the graph and
tenant configuration, constructs decision variables, objective
functions, and constraints, and returns reproducible results. The solver
may be OR-Tools, Gurobi, CPLEX, or another deterministic solver
depending on problem scale and licensing. In the deployments described
in this paper, OR-Tools serves as the primary optimization engine. The
architecture does not claim novelty in optimization algorithms. The
contribution is the integration pattern: the LLM layer helps identify
the problem and explain the result; the graph supplies structured facts;
the solver computes the answer. LLMs decide what should be computed;
deterministic engines compute it.

This separation reduces a major class of errors. In an LLM-only system,
a model may infer a payment recommendation and calculate totals in
generated text. In the three-engine architecture, the model may classify
the request as payment optimization, identify relevant constraints, and
ask the solver to schedule payments. The solver output becomes a
structured artifact. The response agent explains the solver output, and
the validation agent checks that the explanation matches the graph and
solver result. The recommendation is therefore grounded in computation
rather than prose.

The third engine is a multi-agent LLM layer. A single agent that
retrieves data, reasons, calculates, drafts, and validates is difficult
to control. The architecture instead separates responsibilities across
specialized agents. A query analysis agent classifies intent, identifies
entities, detects implicit time windows, and decides which engines are
required. A context assembly agent retrieves graph context, selects
traversal depth, applies filters, and prepares model-readable evidence.
An optimization invocation agent formulates solver inputs when the query
requires computation. A response generation agent produces the
user-facing explanation. A validation agent checks claims against
structured ground truth before the response is returned.

This multi-agent structure matters because financial validation is not
the same as asking another model whether an answer sounds correct.
Validation must compare generated claims to source records and solver
outputs. If the response says that Vendor A has 50,000 outstanding, the
validation agent should verify that Vendor A exists, that the selected
invoices total 50,000, that the invoices are open, that the amounts are
current for the relevant snapshot, and that any cited discount terms
exist. If the response says that paying Vendor A improves discount
capture by 1,000, the validation agent should verify that the solver
output supports that claim. Unsupported claims should be removed,
revised, or escalated for human review.

\begin{figure}[htbp]
\centering
\includegraphics[width=\linewidth]{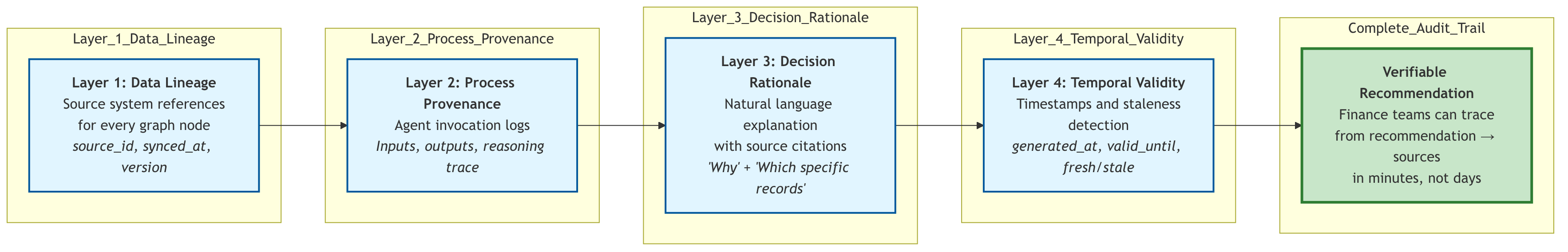}
\caption{Four-layer audit trail architecture: data lineage, process provenance, decision rationale, and temporal validity.}
\label{fig:audit-trail}
\end{figure}

Auditability is built across all three engines. The graph records data
lineage. Solver runs record the problem formulation, input snapshot,
parameters, objective, constraints, and output. Agent logs record query
classification, context selection, tool calls, validation failures, and
final response. A complete audit trail for a payment recommendation may
include the user query, detected time window, invoices retrieved, cash
balance source, tenant policy, solver formulation, selected payment
schedule, validation checks, and response citations. This layered audit
trail is necessary for finance users to trust and challenge the system.

The architecture also supports historical queries. Because the graph
persists historical transaction state, users can ask questions about
trends, causes, and previous decisions. ``Why did aging worsen in
March?'' is not a current-state lookup. It requires comparing aging
buckets over time, identifying invoices that moved between buckets,
checking payment delays, and connecting changes to customers, disputes,
or operational events. A snapshot-only architecture cannot answer this
well. A temporal graph can supply the evidence, and LLM agents can
assemble a narrative explanation validated against historical records.

The coordination challenge is where the system becomes difficult. The
architecture has multiple engines, but a user does not ask, ``Please
traverse the graph to depth three, then run an integer program, then
validate output fields.'' The user asks a business question. The system
must determine whether the question is a lookup, explanation, analysis,
optimization, scenario, or hybrid. It must decide which subgraph to
retrieve, how much history to include, when to invoke the solver, which
tenant rules apply, and how to allocate context window budget. These
decisions are context engineering decisions.

In practice, the coordination layer performs several functions. It
classifies queries. It maps natural language entities to graph entities.
It selects graph traversal strategies. It applies temporal and access
filters. It chooses whether deterministic computation is required. It
prepares solver inputs. It allocates context across retrieved evidence,
calculations, validation traces, and response templates. It routes
outputs to validation. It preserves audit records. Without this layer,
the three engines are merely tools. With it, they become a coherent
financial AI system.

The architecture is intentionally model-agnostic. It does not depend on
a specific LLM vendor or model generation. The production deployments
described in this paper primarily use Gemini as the LLM provider; the
architecture has also been evaluated against other open-source models.
The model-agnostic claim is supported by the fact that neither
correctness of financial facts nor reproducibility of calculations
depends on model behavior---those properties are enforced by the graph,
solver, and validation layer. Better models may improve query
interpretation and explanation quality, but they do not remove the need
for structured source truth, deterministic computation, validation, and
audit trails. This point is important because model capabilities are
improving rapidly. A robust architecture should benefit from stronger
models without becoming dependent on model behavior for facts or
arithmetic.

The architecture is also ERP-agnostic in principle. ERP integrations
normalize source records into graph abstractions. Tenants differ in
chart-of-account conventions, vendor naming practices, tax regimes,
document completeness, and workflow policies. The graph and context
templates absorb those differences so that the reasoning layer can
operate over a consistent conceptual model while preserving source
provenance. This approach does not eliminate integration work, but it
makes cross-tenant learning possible.

An end-to-end flow clarifies the architecture. A user asks, ``Can we pay
the critical vendors due this week without touching payroll reserve?''
The query analysis agent classifies the query as a constrained payment
optimization request. It extracts the time window, recognizes ``critical
vendors'' as a tenant-specific policy category, and identifies ``payroll
reserve'' as a liquidity constraint. The context assembly agent
retrieves open invoices due this week, vendor criticality scores,
discount windows, bank balances, payroll reserve policy, tax
obligations, and recent payment holds. The optimization invocation agent
constructs variables for invoice payment decisions, constraints for
available cash and reserve preservation, and an objective that balances
overdue risk, discount capture, and vendor criticality. The solver
returns a feasible schedule or explains infeasibility. The response
agent turns the schedule into a business explanation. The validation
agent checks that every amount, vendor, invoice, due date, and
constraint in the explanation is supported by the graph or solver
output.

This flow makes two design choices visible. First, the LLM layer is
allowed to interpret intent and translate between human language and
system operations, but it is not allowed to invent the financial state.
Second, the output is not treated as complete until it has passed
validation. The architecture therefore converts an unconstrained
language model into a controlled participant in a larger system. The
model remains useful because it can understand the user's question and
produce a clear explanation; it becomes safer because facts and
calculations are externalized.

The architecture also supports progressive disclosure. Not every user
needs the full audit trace in the first response. A controller may want
a concise answer, while an auditor may want record-level evidence. The
system can present a recommendation with summarized rationale and allow
drill-down into source invoices, solver assumptions, validation checks,
and ERP sync metadata. This is important for usability. Auditability
should be available without overwhelming routine users. The graph and
logs preserve the evidence; the LLM layer can expose the right level of
detail for the user and workflow.

The next section focuses on the coordination layer itself. Context
engineering is the least standardized part of the architecture and the
part most responsible for whether the system behaves like a trustworthy
financial assistant or a fluent but unsafe chatbot.

\section{4. Context Engineering}\label{context-engineering}

\begin{figure}[htbp]
\centering
\includegraphics[width=\linewidth]{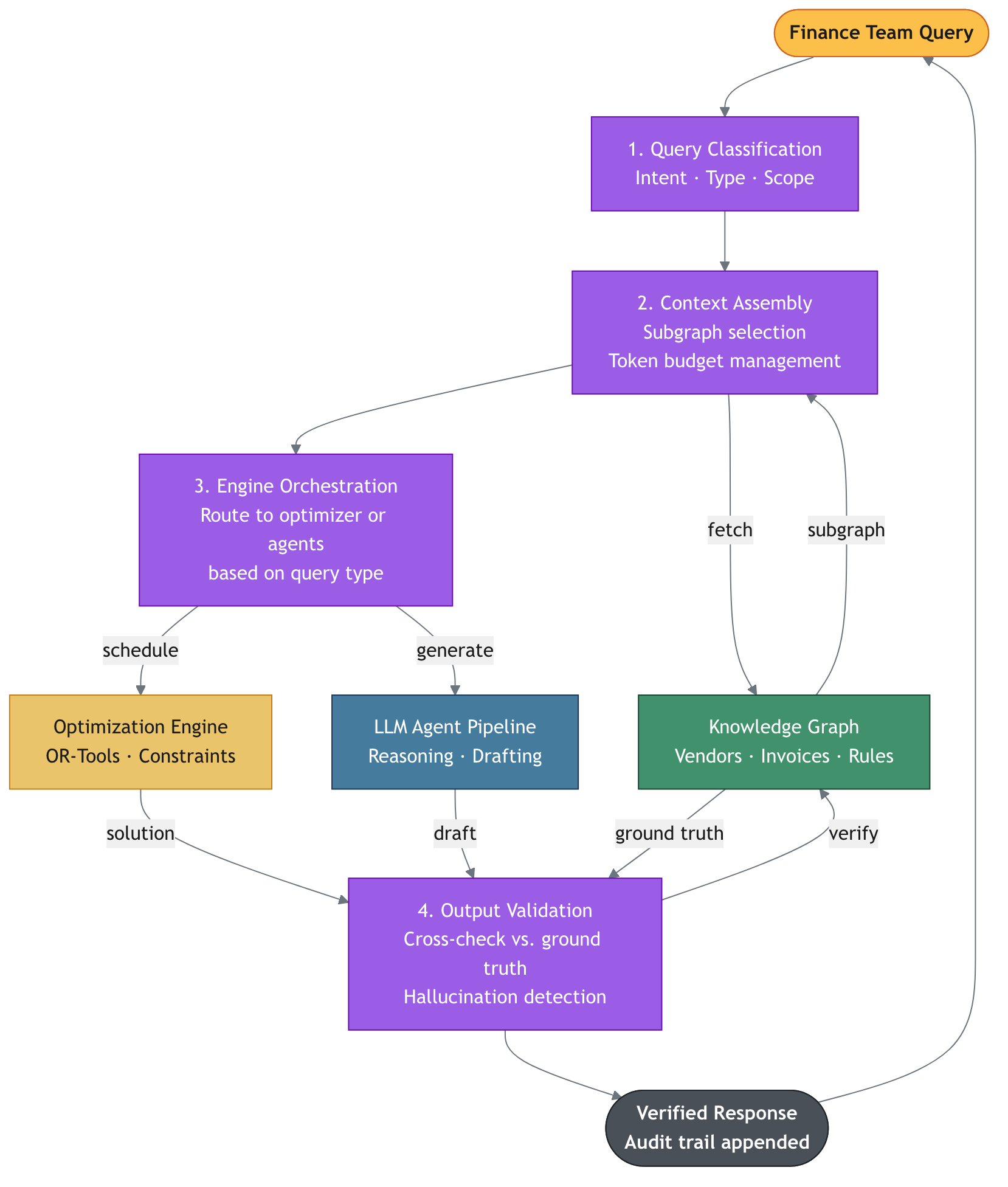}
\caption{Context engineering framework for financial AI. The coordination layer classifies queries, assembles graph context, orchestrates agents and deterministic engines, validates outputs, and manages context budgets.}
\label{fig:context-engineering}
\end{figure}

Context engineering is the central concept of this paper. We
characterize financial context engineering as the process of selecting,
ranking, composing, validating, and budgeting the information required
for a financial AI system to answer a query accurately,
deterministically where needed, and with traceability to source records.
More precisely: given a user query, a tenant configuration, a financial
knowledge graph, available computational engines, applicable policy
constraints, and a context window budget, the context engineering layer
produces a context package and execution plan that maximize
decision-relevant information subject to accuracy, completeness,
recency, access-control, and budget constraints. This characterization
captures the multi-dimensional nature of the problem. A rigorous
optimization formulation would require defining relevance functions and
feasibility constraints that vary by tenant, query type, and data
state---an open research problem we discuss in Section 8.

This definition intentionally includes more than retrieval. A context
package may include graph records, relationship paths, aggregate
metrics, historical snapshots, source citations, solver outputs, tenant
policy parameters, validation results, and response constraints. An
execution plan may specify that the system should first resolve
entities, then retrieve open invoices, then run an optimization model,
then validate the generated explanation against selected invoices and
solver results. Context engineering is therefore both data selection and
workflow selection.

The term context engineering has appeared in prior AI systems
literature, primarily in discussions of agent design, prompt
construction, and context window management. Our usage builds on that
framing but applies it in a domain-specific sense. In financial AI,
context engineering is specifically concerned with structured
relationship retrieval over multi-tenant graphs, temporal validity
enforcement, deterministic computation grounding, access-controlled
evidence assembly, and audit-trail preservation. This domain
specificity---driven by requirements for determinism, multi-tenant
heterogeneity, and regulatory auditability---distinguishes financial
context engineering from general context management in language model
applications.

The distinction from standard RAG is fundamental. RAG systems usually
retrieve text chunks based on semantic similarity to a query. This is
useful for many knowledge tasks, but financial decision tasks require
structured relevance. A semantically similar invoice is not relevant if
it is already paid, belongs to another legal entity, falls outside the
time window, or is inaccessible to the user. A small invoice with a
discount expiring tomorrow may be more decision-relevant than a large
invoice that is not yet due. A vendor relationship may matter because of
criticality rather than textual similarity. Relevance is
multi-dimensional: amount, urgency, compliance risk, relationship
importance, operational dependency, temporal state, and policy priority.

Context engineering begins with query understanding. The system must
determine what type of question has been asked. Lookup queries request
facts: ``What is outstanding for Vendor X?'' Explanation queries ask why
something happened: ``Why did aging increase?'' Optimization queries ask
what action to take under constraints: ``Which invoices should we pay
with 200,000 available cash?'' Scenario queries ask what changes under
assumptions: ``What if Customer Y pays ten days late?'' Hybrid queries
combine several types: ``Which vendors should we pay first and why?''
Correct classification determines engine routing. A lookup may require
graph traversal only. An optimization query requires graph retrieval
plus solver invocation. A scenario may require historical context,
assumptions, and recalculation.

Entity resolution is the next challenge. Finance data contains aliases,
abbreviations, spelling differences, merged vendors, subsidiaries, and
reused names. A user may say ``ABC Supplies'' while the ERP contains
``A.B.C. Industrial Supply Pvt Ltd'' and the payment system contains
``ABC Ind Supplies.'' The context layer must resolve the intended entity
and preserve uncertainty when resolution is ambiguous. It should not
silently choose the wrong vendor. Entity resolution must also respect
tenant scope and access control. A user in one entity should not receive
data from another legal entity unless authorized.

Subgraph selection determines which relationships to include. Too little
context causes omissions; too much context consumes model budget and
increases distraction. For vendor payment, useful subgraphs may include
open invoices, payment terms, discount windows, purchase orders, goods
receipts, dispute flags, vendor criticality, payment history, and cash
constraints. For customer collections, useful subgraphs may include
aging, communications, dispute status, promised payment dates, account
value, service delivery, and credit limits. For tax review, useful
subgraphs may include invoice classifications, tax codes, jurisdiction,
filing period, reconciliation status, and document completeness. Each
query requires a different traversal pattern.

Temporal filtering is essential. Financial facts are time-bound. An
invoice that was open last week may be paid today. A discount window may
have expired. A tax rate may have changed. A cash balance may be a
snapshot from yesterday's bank sync. Context engineering must attach
temporal validity to retrieved facts and ensure that the response
reflects the relevant time window. A question about ``this week'' needs
current obligations and expected cash flows. A question about ``why did
March aging worsen'' needs historical state from March and comparison
periods. Without temporal discipline, the system may mix facts from
different states and produce incorrect explanations.

Tenant adaptation is another defining challenge. The same phrase can
mean different things across industries and organizations. ``Outstanding
liability'' in a manufacturing tenant may require purchase orders, goods
receipts, inventory status, supplier invoices, taxes, and accruals. In a
SaaS tenant it may require subscriptions, deferred revenue, receivables,
churn risk, and customer contracts. In a services tenant it may require
projects, milestones, timesheets, retainers, and unbilled work. In a
trading business it may require inventory financing, supplier credit,
customer advances, and currency exposure. Context templates therefore
need industry and tenant parameters. A universal prompt cannot encode
all these distinctions reliably.

Budget management is a practical constraint. Even large context windows
are finite, and more context is not always better. Context engineering
must allocate space among user query, resolved entities, graph facts,
historical summaries, solver inputs, solver outputs, validation
evidence, policy constraints, and response instructions. The context
should be sufficient and minimal. Sufficient means the answer has the
evidence needed to be accurate and auditable. Minimal means irrelevant
facts are excluded so the model is less likely to attend to noise. This
creates a resource allocation problem similar in spirit to optimization:
choose the most decision-relevant context under a fixed budget.

\begin{figure}[htbp]
\centering
\includegraphics[width=\linewidth]{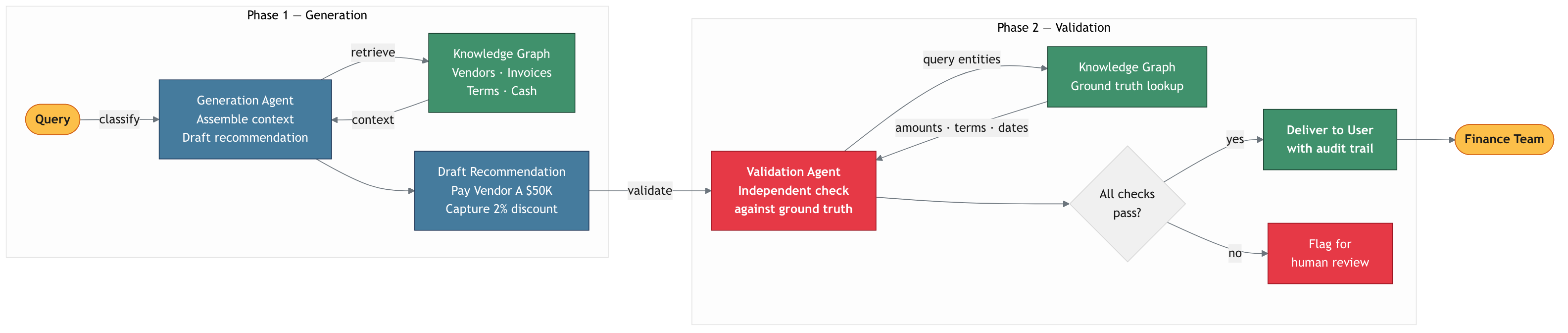}
\caption{Multi-agent validation pattern. Separate agents cross-check LLM outputs against graph ground truth and deterministic solver results before responses reach finance users.}
\label{fig:multi-agent-validation}
\end{figure}

Validation closes the loop. A response should not be returned merely
because it is fluent. Claims must be checked. Entity claims should map
to graph entities. Numerical claims should match source records or
solver outputs. Temporal claims should be valid for the stated period.
Feasibility claims should respect constraints. Source citations should
point to records that support the statement. Unsupported claims should
be rejected. If validation fails, the system may revise the response,
retrieve additional context, ask a clarification question, or route the
case for human review.

An example illustrates the full flow. Suppose a user asks, ``Which
vendors should we prioritize for payment this week if we keep 100,000
reserved for payroll?'' Query understanding classifies the request as
payment optimization with an explicit cash reserve. Entity scope
identifies the relevant legal entities and bank accounts. Subgraph
selection retrieves open vendor invoices, due dates, terms, discounts,
vendor criticality, purchase order status, tax obligations, and current
cash balance. Temporal filtering ensures invoices and cash balances are
current. Tenant policy adds constraints such as mandatory payroll
reserve, strategic supplier priority, and tax-payment deadlines. The
optimization engine computes a schedule. The response agent explains the
recommendation. The validation agent checks that every cited vendor,
invoice, amount, discount, and constraint appears in the graph or solver
output. The audit trail records the full chain.

This example also shows why context engineering is not a prompt trick.
It is a system function that touches data modeling, retrieval,
optimization, validation, access control, and user experience. It
requires domain knowledge because the system must know which facts
matter for which financial decisions. It requires engineering discipline
because the context package must be reproducible and testable. It
requires evaluation methods because the field needs to measure not only
answer quality, but context quality.

Several open problems follow. How should context sufficiency be
measured? How should systems detect missing critical context before
generating an answer? How should context policies adapt across tenants
without leaking data? How should context selection be benchmarked when
production financial data is private? How should graph traversal depth
be optimized for accuracy and cost? How should validation failures be
categorized and learned from? These questions are central to making
financial AI reliable.

The production deployments discussed later suggest that context
engineering becomes a reusable asset. Tenant-specific mappings, industry
templates, validation checks, and traversal patterns accumulate over
time. This accumulation explains why deployment time decreased across
later tenants. The architecture does not become effortless, but repeated
patterns become codified. Context engineering, in this sense, is both a
runtime function and an organizational learning process.

A useful benchmark for context engineering would separate at least four
layers. The first layer would test entity and relationship retrieval:
given a query and a graph, did the system retrieve the records that a
financial expert would need? The second layer would test execution
planning: did the system choose graph lookup, deterministic calculation,
optimization, scenario replay, or human escalation correctly? The third
layer would test claim grounding: does each statement in the final
answer map to a source record, solver output, or explicit assumption?
The fourth layer would test context economy: did the system answer with
the smallest sufficient context, or did it depend on large, noisy
context dumps? Such benchmarks would move the field away from
surface-level answer grading and toward the actual reliability problem.

\section{5. Integration Challenges}\label{integration-challenges}

The three-engine architecture creates a powerful division of labor, but
integration is difficult. The main challenges are boundary setting,
validation, tenant adaptation, reproducibility, and model-agnostic
design.

The first challenge is drawing a boundary between reasoning and
computation. LLMs are useful for interpreting language, detecting
intent, identifying relevant entities, proposing problem structure, and
explaining trade-offs. They are not reliable as the final authority for
arithmetic or optimization. The practical rule is simple: keep
arithmetic and optimization out of the language model and in
deterministic engines. This boundary must be enforced in code, not only
in prompting. If a response contains a total, the total should come from
a calculation or source record. If a response recommends a schedule, the
schedule should come from the optimization engine. If a response cites a
discount, the discount should come from the graph.

The second challenge is validation. Early single-agent prototypes
exposed the risk clearly. A single agent could produce a plausible
recommendation that referenced a nonexistent invoice, attributed a
discount to the wrong vendor, summed invoices incorrectly, or suggested
a payment that violated a cash constraint. The answer sounded credible
because it used finance language, but it was not grounded. Multi-agent
validation reduces this risk by separating generation from checking. The
generation agent drafts an explanation; the validation agent compares
claims against graph records and solver outputs. The output is only safe
if the validation layer has authority to block or revise unsupported
claims.

Validation must be specific. Entity validation checks whether named
vendors, customers, invoices, and accounts exist in the graph and match
the intended tenant. Numerical validation checks totals, dates,
percentages, balances, and calculated metrics. Temporal validation
checks whether the facts were current for the stated period. Constraint
validation checks whether recommendations respect cash limits, due
dates, reserves, tax deadlines, and policy rules. Source validation
checks whether cited records support the text. Feasibility validation
checks whether the recommended action can actually be executed in the
financial workflow. A general ``does this look right?'' model critique
is not enough.

The third challenge is tenant adaptation. Tenants differ at schema,
template, and instance levels. Schema differences include
industry-specific entities and relationships. A manufacturing tenant may
need purchase orders, goods receipts, inventory, supplier criticality,
and production dependencies. A SaaS tenant may need subscriptions,
deferred revenue, churn risk, and recurring receivables. A services
tenant may need project milestones, timesheets, retentions, and unbilled
work. Template differences include compliance regimes such as GST, VAT,
sales tax, and local filing calendars. Instance differences include
chart-of-account conventions, payment policies, approval thresholds,
naming practices, and historical behavior.

Tenant adaptation creates a tension between reuse and customization. A
fully custom implementation for every tenant is slow and expensive. A
fully generic implementation is shallow and often wrong. The production
pattern is parameterized reuse: common graph primitives, common ERP
mappings, common validation checks, and common agent workflows, with
tenant-specific context templates and policy parameters. Over time,
these templates become more precise. Deployment time falls because the
system learns reusable patterns, but each tenant still requires
data-quality work and policy calibration.

The fourth challenge is reproducibility. Financial AI outputs can change
because source data changes, graph mappings change, tenant policies
change, solver versions change, prompt templates change, or model
versions change. Some change is correct: if an invoice is paid, the
answer should update. But uncontrolled change undermines trust. The
architecture therefore needs versioned templates, graph snapshots or
synchronization timestamps, solver configuration logs, model
identifiers, prompt versions, and validation traces. Reproducibility is
defined relative to fixed data, fixed templates, fixed solver settings,
and fixed model versions. Without these controls, users cannot
distinguish legitimate updates from system drift.

The fifth challenge is model-agnostic design. Model capabilities will
improve. Future LLMs may classify queries better, follow tool contracts
more reliably, and generate clearer explanations. But financial
requirements will not disappear. A better model still needs source
truth. It still should not invent invoices. It still should not perform
final payment optimization in prose. It still needs audit trails and
access control. The architecture should use models where they add value
while constraining them away from roles where deterministic systems are
superior.

Integration also creates operational challenges. Latency must be managed
because graph traversal, solver invocation, validation, and generation
can be expensive. Error handling must be explicit because each engine
can fail differently: ERP sync may be stale, graph resolution may be
ambiguous, solver may find no feasible solution, validation may reject a
claim, or the model may produce an incomplete explanation. The system
needs fallback behaviors. It may ask the user for clarification, return
a partial answer with caveats, escalate to human review, or provide a
diagnostic explanation of why no recommendation can be made.

Security and privacy must be integrated into context assembly. Context
should be filtered by user permission before it reaches the model.
Multi-tenant deployments must prevent cross-tenant leakage. Audit logs
may contain sensitive information and need retention policies,
encryption, and access controls. Context engineering cannot be separated
from security because the selected context is exactly what the model
sees and what the logs preserve.

Finally, integration requires product discipline. Finance users need
answers that are actionable, not merely technically correct. A payment
recommendation should include the recommended action, rationale,
constraints, expected benefit, risks, and source links. A cash-flow
explanation should identify drivers, quantify impact, and separate facts
from assumptions. A tax-risk review should identify records requiring
human attention. The architecture must therefore connect technical
grounding to usable workflows.

The central integration lesson is that reliability emerges from explicit
responsibility boundaries. When engine roles blur, hallucination and
audit risk increase. When they are enforced, the system becomes easier
to test, debug, and trust.

Another integration challenge is failure communication. In many software
systems, failure can be expressed as an error code. In financial AI, a
failed recommendation often still contains useful information. If a
solver finds no feasible payment schedule under the requested cash
reserve, the system should explain which constraints make the request
infeasible and what alternatives are available. If entity resolution is
ambiguous, the system should show the candidate vendors and ask the user
to choose. If ERP sync is stale, the system should state the last
synchronization timestamp and avoid pretending the answer is current.
Clear failure communication is part of trust. It turns system
limitations into actionable workflow signals rather than hidden risks.

Monitoring must also be domain-specific. Generic latency and uptime
metrics are not enough. The system should monitor validation failure
categories, entity-resolution ambiguity, stale-data usage, solver
infeasibility, unsupported-claim attempts, source-system sync delays,
and user overrides. These metrics reveal whether the architecture is
improving or merely generating plausible responses. For example, a drop
in user-visible hallucinations accompanied by a rise in validation
failures may indicate that validation is working but upstream context
assembly needs improvement. A rise in entity-resolution ambiguity may
indicate data-quality drift in an ERP integration. Observability is
therefore part of the financial AI control system.

\section{6. Production Deployment}\label{production-deployment}

\begin{figure}[htbp]
\centering
\includegraphics[width=\linewidth]{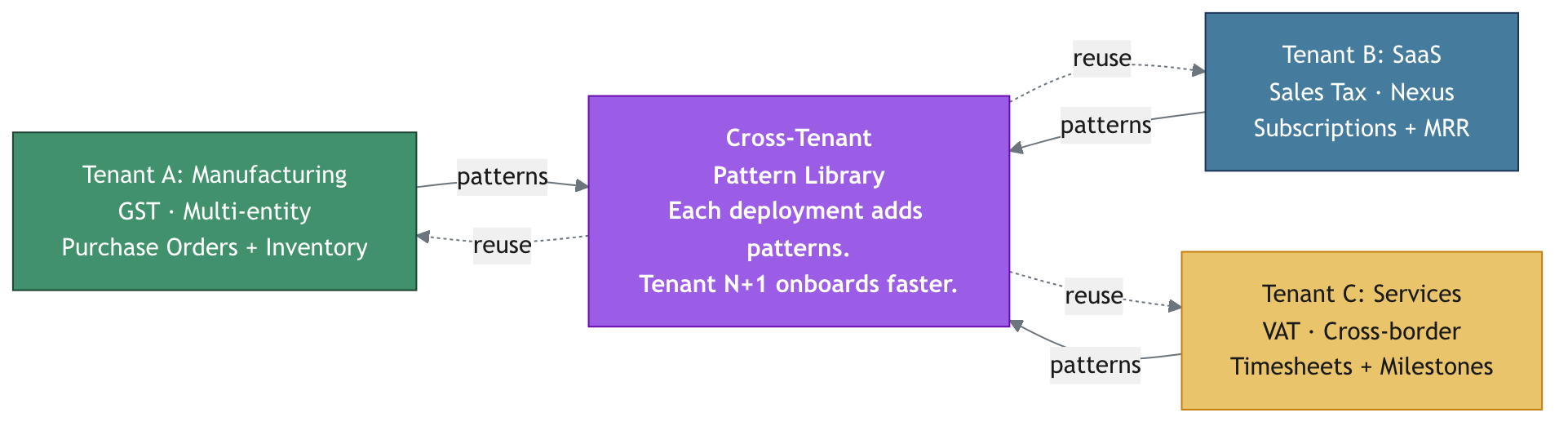}
\caption{Representative production patterns for tenant adaptation, deployment learning, and cross-tenant pattern reuse.}
\label{fig:production-patterns}
\end{figure}

The architecture has been deployed across 15 production deployments
selected from a 150+ client base over approximately 11 months.
Deployments were selected based on deployment maturity, data
availability, and organizational data readiness; this selection
criterion means results may overrepresent customers with relatively
mature data practices. The deployments span manufacturing, SaaS,
services, e-commerce, and trading businesses. ERP integrations include
Zoho, Tally, QuickBooks, and Acumatica. Tenants process approximately 50
to 500 invoices per optimization query, with graph sizes ranging from
tens of thousands to more than one million nodes depending on tenant
complexity, historical depth, and integration scope.

The observations in this section should be interpreted as production
evidence rather than controlled experimental proof. The deployments
occurred in real customer settings with changing data, improving
tooling, evolving templates, and different tenant mixes. The results are
therefore useful for architectural learning, but not sufficient for
causal claims. All observations are reported on an aggregated basis;
individual customer names, transaction records, and proprietary data are
not disclosed.

The first observation is that single-agent LLM pipelines produced
unacceptable financial hallucinations in early pilots. The errors were
not abstract. They included fabricated invoice identifiers, invented
payment discounts, nonexistent vendors, incorrect invoice totals, stale
due dates, and recommendations that ignored cash constraints. These
errors often appeared in otherwise polished explanations. This made them
particularly dangerous because users could be persuaded by fluency
unless the system exposed source records.

To measure hallucination rates, we evaluated a labeled subset of
approximately 500 AI-generated recommendations against ground-truth ERP
records. Evaluation was conducted by both engineers and finance users,
supported by automated validation against known-correct datasets. A
response was classified as containing a hallucination if any
claim---vendor identity, invoice amount, payment term, tax code, or
computed total---was not supported by source records. In pilot usage
with single-agent pipelines, approximately 36 to 43 of 500 evaluated
responses contained at least one unsupported claim, yielding a rate of
approximately 7--9 percent. The rate was higher for optimization queries,
where the model was more likely to generate specific totals without
grounding them in the solver output.

The move to graph-backed multi-agent validation changed the failure
profile. With structured grounding and validation, generated claims were
checked against graph records and solver outputs before being returned.
In the equivalent labeled review of production recommendations,
unsupported claims reaching users fell to approximately 4 to 5 per 500
responses, below 1 percent. Remaining issues were more often validation
failures, missing context, or data-quality exceptions than unsupported
claims reaching users. This should not be interpreted as a universal
guarantee. The reduction is observational and confounded by system
maturity, better templates, and deployment learning. Still, the
production pattern supports the architectural claim that validation
against structured ground truth is necessary for financial AI.

The second observation is that ERP data quality consumed more effort
than AI modeling. Real financial data is messy. Vendor names differ
across systems. Duplicate vendors exist. Payment terms are stored as
free text. Dates and currencies are inconsistent. Tax codes may be
missing or overloaded. Source records may be incomplete. Historical
corrections may not be synchronized cleanly. User-defined fields may
contain critical business logic. Before the AI system can reason, the
data must be normalized, mapped, deduplicated, and enriched. Entity
resolution and source mapping became core engineering tasks rather than
preprocessing details.

This data-quality work directly affects context engineering. If vendor
identity is ambiguous, context retrieval may assemble the wrong
invoices. If payment terms are free text, discount windows may be
missed. If tax codes are inconsistent, compliance context may be
incomplete. If synchronization timestamps are not tracked, temporal
validity is unclear. The system can detect some inconsistencies, but it
cannot manufacture missing source truth. Production reliability
therefore depends on data contracts between ERP integration, graph
construction, context assembly, and validation.

The third observation is that industry context matters. The same
question requires different evidence in different industries. A
manufacturing tenant asking about payables may require supplier
criticality, inventory impact, purchase orders, goods receipts, and
production schedules. A SaaS tenant may require customer subscriptions,
deferred revenue, renewal dates, churn risk, and receivables. A services
tenant may require project milestones, timesheets, client approvals,
retainers, and unbilled work. An e-commerce tenant may require inventory
velocity, marketplace settlements, refunds, and payment gateway timing.
A trading business may require currency exposure, supplier credit,
customer advances, and shipment status. Context engineering must
therefore be industry-aware.

The fourth observation is that deployment time decreased as reusable
patterns accumulated. Early tenants took roughly six weeks from project
kickoff to customer production handover. Later tenants with reusable ERP
mappings, graph templates, validation checks, and industry-specific
context policies averaged approximately two weeks from kickoff to
handover, though this varied by tenant complexity, data maturity, and
ERP integration scope. This improvement reflects learning across
tenants, but also improved tooling, better implementation processes, and
possibly easier customer selection. The important architectural lesson
is that context engineering assets are reusable. A validated mapping for
a common ERP object, a payment-term parser, an entity-resolution
pattern, or an industry context template can reduce future
implementation effort.

The fifth observation is that users valued explanation and traceability,
not only recommendations. Finance teams often asked why a recommendation
was made, which records supported it, what assumptions were used, and
what alternatives were considered. A recommendation without explanation
was treated as an opaque model output. A recommendation with
source-linked rationale became a decision-support artifact. This
reinforced the importance of audit trails and response design. The
system must show the path from source data to recommendation, especially
when the recommendation affects cash, vendor relationships, or
compliance.

The sixth observation is that human review remains part of the workflow.
The goal is not to remove finance judgment. The goal is to reduce manual
analysis, surface better options, and make decisions easier to justify.
Certain cases should be escalated: ambiguous entity resolution, missing
source records, infeasible optimization, large-value payments,
compliance uncertainty, or validation failures. Human-in-the-loop design
is therefore not a weakness. It is a control mechanism appropriate for
high-stakes domains.

The seventh observation is that context policies became more important
than prompt wording. Prompt improvements helped, but the largest
reliability gains came from better retrieval, graph structure,
validation, and tenant-specific context assembly. This supports the
paper's core claim. Financial AI quality depends less on writing a
clever universal prompt and more on constructing the right context and
execution path for each query.

Several deployment metrics were useful. Hallucination rate measured
unsupported claims reaching the user. Validation failure rate measured
claims blocked or revised before response. Entity resolution accuracy
measured correct mapping from user language to graph entities. Solver
feasibility rate measured whether optimization queries produced valid
solutions. Context sufficiency measured whether retrieved evidence
supported the answer without manual follow-up, although this metric
remains less standardized. Time-to-go-live measured implementation
efficiency. User adoption measured whether finance teams continued using
the system for recurring workflows.

The production experience also revealed what not to overclaim. The
deployments do not prove that the architecture is optimal. They do not
prove that hallucination rates will remain below 1\% across all tenants,
query types, and future models. They do not isolate the causal effect of
each component. They do not provide a public reproducible benchmark
because customer data is private. The contribution is instead an
empirically motivated architectural pattern and a set of lessons from
real deployments.

The most useful deployment reviews were trace-based. Instead of only
inspecting final responses, reviewers inspected the complete trace: user
query, query classification, entity resolution, retrieved graph context,
solver inputs, solver output, generated explanation, validation checks,
and user feedback. Trace review made errors actionable. If the wrong
vendor was selected, the issue was entity resolution. If a relevant
invoice was missing, the issue was traversal or temporal filtering. If
the solver result was correct but the explanation overstated certainty,
the issue was response generation. If validation passed an unsupported
claim, the issue was validation policy. This trace-based review process
is an important practical method for improving multi-engine financial AI
systems.

The most important lesson is that reliability is systemic. No single
component solved the problem. Better prompts did not solve it alone.
Better models did not solve it alone. Better graph schemas did not solve
it alone. The improvement came from aligning source data, graph
structure, deterministic computation, agent roles, validation, audit
logs, and context policies. Financial AI is an engineered system, not a
model call.

\section{7. Related Work}\label{related-work}

This work sits at the intersection of LLM-based financial analysis,
knowledge graphs, mathematical optimization, multi-agent systems,
retrieval-augmented generation, and AI system design. Its contribution
is architectural and empirical rather than algorithmic: it integrates
existing classes of engines into a production-validated pattern for
financial decision support and identifies context engineering as the
coordination problem.

LLM research established the foundation for natural language interfaces.
Transformer architectures and large-scale pretraining enabled models
that can parse user questions, summarize evidence, generate
explanations, and participate in multi-step workflows {[}Vaswani et al.,
2017; Brown et al., 2020{]}. Financial LLM work, including
domain-specialized models such as BloombergGPT, shows the value of
finance-specific corpora and terminology {[}Wu et al., 2023b{]}.
However, domain fluency does not by itself solve grounding, determinism,
or auditability. A model trained on financial text may understand terms
such as aging, working capital, and payment discount, but it still needs
current tenant-specific source data and deterministic tools.

Work on hallucination and non-determinism explains why LLM-generated
text cannot be treated as a financial source of truth {[}Ji et al.,
2023; Atil et al., 2024{]}. Hallucination is often framed as factual
error in generated content. In finance, hallucination must be
operationalized as unsupported or incorrect claims about entities,
amounts, dates, terms, calculations, or constraints. This paper's
validation architecture addresses hallucination by comparing generated
claims against structured graph data and solver outputs. The focus is
prevention and containment through architecture, not post-hoc textual
critique alone.

Financial knowledge graphs and ontologies provide another foundation.
Standards such as FIBO {[}EDM Council, 2023{]} demonstrate the value of
representing financial entities and relationships semantically.
Knowledge graphs support relationship traversal, lineage, and reasoning
over connected data. Existing financial knowledge graphs, however, are
often designed for formal semantics, human query, or enterprise
integration rather than LLM context assembly. This work emphasizes
LLM-aware graph design: relationship labels, node summaries, provenance
fields, and hierarchical structures that make retrieved context
interpretable by agents while preserving source traceability.

Mathematical optimization has a long history in finance and operations
research. Portfolio optimization, cash-flow planning, payment
scheduling, credit allocation, and supply-chain finance all use
deterministic optimization methods {[}Markowitz, 1952; Konno \&
Yamazaki, 1991; Baccarin, 2009; Wuttke et al., 2016{]}. This paper does
not propose a new solver or optimization algorithm. Instead, it shows
how optimization can function as one engine in an AI system. LLMs
formulate and explain; solvers compute. This integration allows finance
users to access optimization through natural language while preserving
reproducibility.

Multi-agent LLM frameworks such as AutoGen and CrewAI provide patterns
for role-based agent collaboration {[}Wu et al., 2023a; CrewAI, 2024{]}.
These frameworks are useful for decomposing tasks and coordinating tool
use. The contribution here is domain-specific: multi-agent validation
against financial ground truth. The validation agent is not merely
another conversational participant. It has a defined responsibility to
check entity existence, numerical accuracy, temporal validity,
feasibility, and source support. This domain-specific grounding is what
makes the multi-agent pattern relevant to finance.

Retrieval-augmented generation improves factuality by supplying external
context to models {[}Lewis et al., 2020{]}. GraphRAG extends retrieval
by using graph structure and community summaries {[}Edge et al.,
2024{]}. Financial context engineering builds on these ideas but differs
in several ways. It retrieves structured records and relationship paths,
not just semantically similar text. It must account for temporal
validity, access control, tenant-specific schemas, deterministic
computation, and validation. It must optimize relevance across financial
dimensions such as urgency, magnitude, compliance risk, relationship
criticality, and operational impact. The output is not only a prompt
context but an auditable execution package.

Practitioner literature on AI system design increasingly emphasizes that
effective applications require orchestration, tools, memory, guardrails,
and feedback loops rather than a standalone model call {[}Anthropic,
2024{]}. Context engineering can be understood as a domain-specific part
of that broader discipline. In financial AI, context engineering is
especially central because the context determines whether the model
operates over real business state or plausible fragments. The field
still lacks formal benchmarks and metrics for context quality in
structured, high-stakes domains.

The closest prior systems combine some of the required pieces but not
all. LLM financial assistants provide natural language interaction but
often lack deterministic optimization and audit trails. Financial
knowledge graph systems provide structured context but often lack
conversational workflows and solver integration. Optimization systems
provide reproducible decisions but require technical formulation and do
not explain themselves naturally. Multi-agent frameworks provide
orchestration but not finance-specific validation. Standard RAG provides
retrieval but not structured, temporal, tenant-aware context
engineering. The three-engine architecture attempts to combine these
capabilities in a way that matches financial requirements.

This positioning also clarifies what the paper does not claim. It does
not claim that knowledge graphs are new, that optimization is new, or
that multi-agent LLMs are new. It claims that reliable financial AI
requires these components to be composed with explicit boundaries and
coordinated through context engineering. The production observations
support this claim by showing where simpler architectures failed and
which patterns improved reliability.

\section{8. Limitations}\label{limitations}

This work has several limitations. The most important is that the
evidence is observational. The deployment observations come from real
production systems, not randomized controlled experiments. Hallucination
reductions, deployment acceleration, and user adoption may be affected
by many confounding factors: system maturity, improved prompts, better
data pipelines, customer selection, team learning, model upgrades, and
process improvements. The reported below-1\% observed hallucination rate
should therefore be interpreted as a production observation under
specific conditions, not a universal benchmark.

The sample is also limited. The deployments cover 15 cases from a 150+
client base; selection based on data maturity means customers with
stronger data foundations are likely overrepresented. The findings may
not transfer directly to microbusinesses, highly regulated banks,
public-sector finance, or enterprises with mature internal data
platforms. The architecture is designed to be general across financial
operations, but the evidence is strongest for the deployment contexts
described.

The system depends heavily on data quality. If ERP data is incomplete,
stale, duplicated, incorrectly mapped, or missing critical fields, the
graph and downstream reasoning inherit those problems. Context
engineering can detect some inconsistencies and validation can block
unsupported claims, but no architecture can fully compensate for missing
source truth. This limitation is practical and fundamental. Financial AI
reliability begins with financial data reliability.

External reproducibility is limited because production financial data is
private. The exact graphs, invoices, payments, vendor records, tax
details, and user queries cannot be released. This makes public
benchmarking difficult. Synthetic datasets, anonymized graph schemas,
replayable validation traces, and task suites are needed for community
evaluation. Without such artifacts, the field risks relying on anecdotal
claims that are hard to compare.

The architecture introduces complexity. A three-engine system is more
complex than a chatbot. It requires ERP integration, graph construction,
solver formulation, agent orchestration, validation logic, audit
storage, and monitoring. This complexity is justified only when the
domain requires reliability, auditability, and deterministic decisions.
For low-stakes summarization or exploratory analysis, a simpler
architecture may be sufficient. The paper's claims are targeted at
operational financial decision support, not every finance-related AI use
case.

Context engineering itself lacks mature theory and metrics. We can
measure final answer correctness, hallucination rate, validation failure
rate, and user adoption. But we lack direct metrics for context
sufficiency, minimality, relevance, freshness, and fairness across
tenants. A system may answer correctly because it retrieved excessive
context, or answer incorrectly because it omitted one critical record.
The field needs methods to evaluate the context package itself, not only
the final response.

Security and privacy remain open challenges. Multi-tenant systems must
prevent leakage of customer-specific data, patterns, and context
templates. Context assembly must enforce user permissions before
information reaches an LLM. Audit logs may contain sensitive financial
information and require encryption, retention policies, access controls,
and deletion workflows. Model providers, deployment environments, and
data residency requirements may impose additional constraints. These
issues are not peripheral; they shape the architecture.

The model-agnostic design also creates evaluation challenges. If the
underlying LLM changes, query classification, tool-use reliability, and
explanation style may change. Versioning can help, but production
systems must monitor behavior across model upgrades. A stronger model
may reduce some failure modes while introducing others. The architecture
is designed to isolate deterministic facts from model behavior, but the
LLM layer still affects user experience and workflow reliability.

Finally, the paper does not provide a full formal specification of all
optimization models, graph schemas, or validation rules. It presents an
architecture and production lessons. Future work should provide open
schemas, reference implementations, synthetic benchmarks, and formal
evaluation protocols---and should quantify economic impact in terms of
time saved, discount capture improvement, and error-prevention value,
metrics that depend on tenant baseline maturity and remain early-stage
here.

\section{9. Conclusion}\label{conclusion}

Financial AI requires more than fluent language generation. Finance
teams need systems that understand business questions, reason over
relationships, compute deterministic recommendations, explain
trade-offs, and preserve audit trails. LLMs are valuable for language
and orchestration, but they are not sufficient as standalone financial
decision engines. Knowledge graphs, optimization solvers, and
rules-based systems each address part of the problem, but none provides
the complete experience alone.

This paper presented a three-engine architecture for reliable financial
AI. A financial knowledge graph provides persistent context,
relationship reasoning, provenance, and temporal state. A deterministic
optimization engine computes constrained decisions such as payment
schedules and cash allocation. A multi-agent LLM layer interprets
queries, assembles tasks, explains outputs, and validates generated
claims against structured ground truth. The engines are coordinated
through context engineering: selecting, ranking, composing, validating,
and budgeting the evidence and execution steps required for each query.

The production experience summarized here suggests that the architecture
addresses real failure modes. Single-agent LLM pipelines produced
unacceptable financial hallucinations. Graph-backed multi-agent
validation reduced observed hallucinations to below 1\% in production
usage. ERP data quality and tenant heterogeneity dominated
implementation effort. Deployment time decreased as reusable mappings,
templates, and validation patterns accumulated. These observations are
not definitive causal proof, but they provide practical evidence for the
architectural direction.

The broader contribution is the framing of context engineering as a
research problem. In high-stakes structured domains, the central
question is not only what a model can generate, but what context it
receives, how that context was selected, how it is validated, and how
the answer can be audited. Financial context engineering requires
structured graph traversal, temporal validity, tenant adaptation,
deterministic tool invocation, access control, and claim validation.
Standard RAG is an important starting point, but it is not enough.

Several research directions follow. The field needs benchmarks for
financial context quality, synthetic but realistic financial graphs,
replayable validation tasks, open evaluation protocols, and metrics for
sufficiency, minimality, freshness, and traceability. It needs methods
for learning cross-tenant patterns without leaking private data. It
needs better ways to test model upgrades in multi-engine systems. It
needs human-in-the-loop workflows that treat escalation as a control
mechanism rather than a failure. It needs security models for context
assembly and audit logs.

The central claim is simple: reliable financial AI is a systems problem.
Models matter, but architecture matters more. The path from source
record to recommendation must be engineered. The graph must preserve
truth. The solver must compute. The agents must communicate and
validate. The context layer must decide what evidence matters. When
these components are aligned, financial AI can become more than a
conversational interface. It can become auditable decision
infrastructure for finance teams.

The practical implication is that teams building financial AI should
invest early in the unglamorous layers: source-system contracts, graph
schema design, entity resolution, temporal metadata, context policies,
validation datasets, and audit logs. These layers are less visible than
a conversational interface, but they determine whether the interface
deserves trust. A stronger model can improve the experience, but it
cannot rescue a system that lacks source truth, deterministic
computation, and validation. The next generation of financial AI should
therefore be judged not only by how well it speaks, but by how well it
knows what it is allowed to say.

\section{References}\label{references}

\begin{itemize}
\tightlist
\item
  Anthropic. (2024). \emph{Building Effective Agents}.
  https://www.anthropic.com/research/building-effective-agents (accessed
  June 2026).
\item
  Atil, B. et al.~(2024). \emph{Non-Determinism of ``Deterministic'' LLM
  Settings}. arXiv:2408.04667.
\item
  Baccarin, S. (2009). \emph{Optimal impulse control for cash management
  with quadratic holding-penalty costs}. Decisions in Economics and
  Finance.
\item
  Brown, T. B. et al.~(2020). \emph{Language Models are Few-Shot
  Learners}. NeurIPS.
\item
  CrewAI. (2024). \emph{CrewAI: Framework for orchestrating role-playing,
  autonomous AI agents}. https://github.com/crewAIInc/crewAI (accessed
  June 2026).
\item
  Edge, D. et al.~(2024). \emph{From Local to Global: A Graph RAG
  Approach to Query-Focused Summarization}. arXiv:2404.16130.
\item
  EDM Council. (2023). \emph{Financial Industry Business Ontology
  (FIBO)}. https://spec.edmcouncil.org/fibo/ (accessed June 2026).
\item
  Ji, Z. et al.~(2023). \emph{Survey of Hallucination in Natural
  Language Generation}. ACM Computing Surveys, 55(12), 1--38.
  https://doi.org/10.1145/3571730.
\item
  Konno, H. and Yamazaki, H. (1991). \emph{Mean-absolute deviation
  portfolio optimization model and its applications to Tokyo stock
  market}. Management Science.
\item
  Lewis, P. et al.~(2020). \emph{Retrieval-Augmented Generation for
  Knowledge-Intensive NLP Tasks}. NeurIPS.
\item
  Markowitz, H. (1952). \emph{Portfolio Selection}. The Journal of
  Finance.
\item
  Vaswani, A. et al.~(2017). \emph{Attention Is All You Need}. NeurIPS.
\item
  Wu, Q. et al.~(2023a). \emph{AutoGen: Enabling Next-Gen LLM
  Applications via Multi-Agent Conversation}. arXiv:2308.08155.
\item
  Wu, S. et al.~(2023b). \emph{BloombergGPT: A Large Language Model for
  Finance}. arXiv:2303.17564.
\item
  Wuttke, D. A. et al.~(2016). \emph{Supply chain finance: Optimal
  introduction and adoption decisions}. International Journal of
  Production Economics.
\end{itemize}

\end{document}